\pdfoutput=1

\documentclass[11pt]{article}

\usepackage{acl}

\usepackage{times}
\usepackage{latexsym}

\usepackage[T1]{fontenc}

\usepackage[utf8]{inputenc}

\usepackage{microtype}

\usepackage{url}

\usepackage{multirow}
\usepackage{multicol}
\usepackage{graphicx}
\usepackage{amsmath}
\usepackage{amssymb}
\usepackage{bbm}
\usepackage{subfigure}
\usepackage{amsfonts}
\usepackage{bm}
\usepackage{bbding}
\usepackage{booktabs}
\usepackage{color}
\usepackage[normalem]{ulem}
\useunder{\uline}{\ul}{}

\usepackage[noabbrev,capitalize]{cleveref} 

\crefname{equation}{equation}{equations}   
\crefname{footnote}{footnote}{footnotes}   
\crefname{line}{line}{lines}               
\crefname{section}{\S}{\S\S}
\Crefname{section}{\S}{\S\S}    
\crefformat{section}{#2\S#1#3}  
\Crefformat{section}{#2\S#1#3}
\crefrangeformat{section}{\S\S#3#1#4--#5#2#6}
\Crefrangeformat{section}{\S\S#3#1#4--#5#2#6}

\usepackage{enumitem}
\setenumerate[1]{itemsep=0pt,partopsep=0pt,parsep=\parskip,topsep=5pt}
\setitemize[1]{itemsep=0pt,partopsep=0pt,parsep=\parskip,topsep=2pt}
\setdescription{itemsep=0pt,partopsep=0pt,parsep=\parskip,topsep=5pt}

\usepackage{makecell}

\title{Dynamic Prefix-Tuning for Generative Template-based Event Extraction}

\author{
Xiao Liu\textsuperscript{1,2,3,4} \and Heyan Huang\textsuperscript{1,2,3,4,}\thanks{\ \ Corresponding author.} \and Ge Shi\textsuperscript{5} \and Bo Wang\textsuperscript{1,2,3,4}\\
\textsuperscript{1}School of Computer Science and Technology, Beijing Institute of Technology \\
\textsuperscript{2}Beijing Engineering Research Center of High Volume Language Information \\Processing and Cloud Computing Applications \\
\textsuperscript{3}Key Laboratory of Intelligent Information Processing and Information Security, \\Ministry of Industry and Information Technology, China \\
\textsuperscript{4}Southeast Academy of Information Technology, Beijing Institute of Technology \\
\textsuperscript{5}Faculty of Information Technology, Beijing University of Technology \\
\texttt{\{xiaoliu,hhy63,bwang\}@bit.edu.cn}, \texttt{shige@bjut.edu.cn} \\
}

\begin{document}
\maketitle
\begin{abstract}

We consider event extraction in a generative manner with template-based conditional generation.
Although there is a rising trend of casting the task of event extraction as a sequence generation problem with prompts, these generation-based methods have two significant challenges, including using suboptimal prompts and static event type information.
In this paper, we propose a generative template-based event extraction method with dynamic prefix (\textsc{GTEE-DynPref}) by integrating context information with type-specific prefixes to learn a context-specific prefix for each context.
Experimental results show that our model achieves competitive results with the state-of-the-art classification-based model \textsc{OneIE} on ACE 2005 and achieves the best performances on ERE.
Additionally, our model is proven to be portable to new types of events effectively.

\end{abstract}

\section{Introduction}

\begin{figure*}[!t]
  \centering
  \setlength{\belowcaptionskip}{-0.3cm}
  \includegraphics[width=\textwidth]{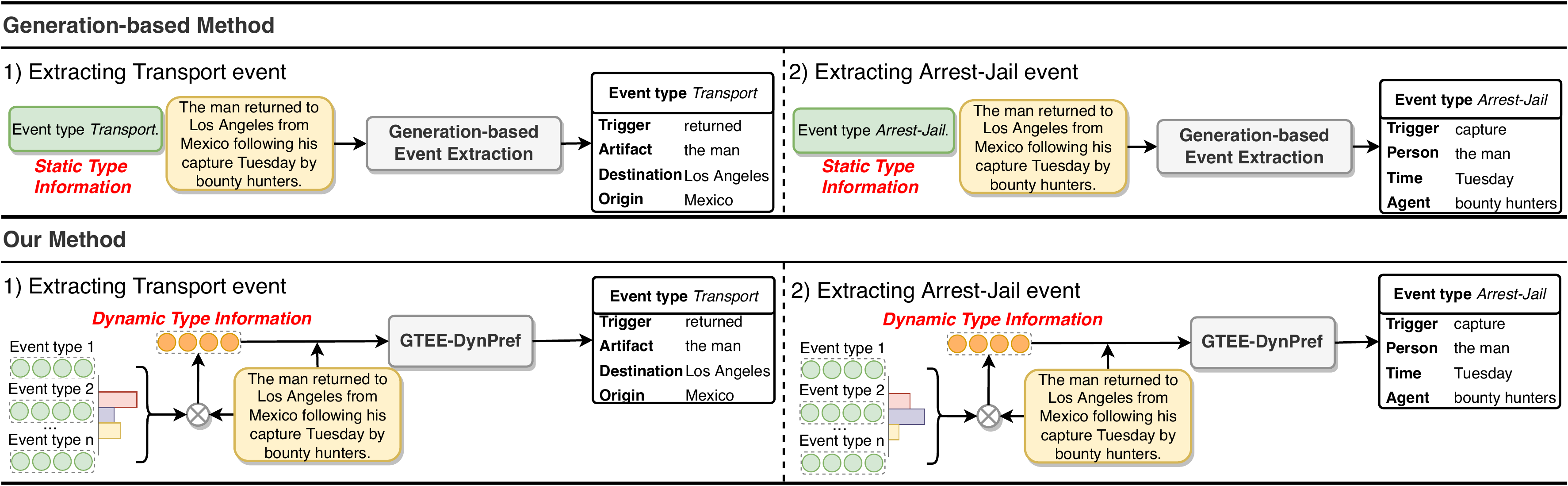}
  \caption{
    Comparision between the generation-based methods and our method \textsc{GTEE-DynPref}.
  }
  \label{fig:introduction}
\end{figure*}

Event extraction is an essential yet challenging task for natural language understanding.
Given a piece of text, event extraction systems need to recognize event triggers with specific types and the event arguments with the correct roles in each event record according to an event ontology, which defines the event types and argument roles \citep{doddington-etal-2004-automatic,ahn-2006-stages}.
As an example, the context in \cref{fig:introduction} contains two event records, a \texttt{Transport} event triggered by ``\textit{returned}'' and an \texttt{Arrest-Jail} event triggered by ``\textit{capture}''.
In the \texttt{Transport} event, the \texttt{Artifact} is ``\textit{the man}'', the \texttt{Destination} is ``\textit{Los Angeles}'' and the \texttt{Origin} is ``\textit{Mexico}''.
In the \texttt{Arrest-Jail} event, the \texttt{Person} is ``\textit{the man}'', the \texttt{Time} is ``\textit{Tuesday}'' and the \texttt{Agent} is ``\textit{bounty hunters}''.
In this work, we focus on the task setting of extracting events without gold entity annotations, which is more practical in real-world applications.

Most of the event extraction work treats the extraction of event triggers and event arguments as several classification tasks, either learned in a pipelined framework \cite{ji-grishman-2008-refining,liu-etal-2020-event, du-cardie-2020-event,li-etal-2020-event} or a joint formulation \cite{li-etal-2013-joint,
yang-mitchell-2016-joint,nguyen-etal-2016-joint-event, liu-etal-2018-jointly, wadden-etal-2019-entity,lin-etal-2020-joint}.

There is a rising trend of casting the task of event extraction as a sequence generation problem by applying special decoding strategies \cite{paolini2021structured,lu-etal-2021-text2event} or steering pretrained language models to output conditional generation sequences with discrete prompts \cite{li-etal-2021-document,DBLP:journals/corr/abs-2108-12724}.
Compared with classification-based methods, this line of work is more data-efficient and flexible, which requires less annotated data to achieve acceptable model performances, being easier to extend to new event types by slightly modifying the designed prompts and decoding strategies.

However, these generation-based methods have two significant challenges, which impede achieving competitive results with the classification-based methods.
(1) \textbf{suboptimal prompts}: First, they manually design prompts for each event type \cite{li-etal-2021-document,DBLP:journals/corr/abs-2108-12724}, which are suboptimal without tuning and largely affect the model performances.
(2) \textbf{static event type information}: Second, when extracting events of a particular type, recent generation-based methods will receive the same event type information concerning only the running event type, regardless of the associations between other possible event types.

To alleviate the above two challenges, we propose a generative template-based event extraction method with dynamic prefixes, denoted as \textsc{GTEE-DynPref}.
As demonstrated in \cref{fig:introduction}, we follow the previous work \cite{li-etal-2021-document,DBLP:journals/corr/abs-2108-12724}, extracting event records one type by one type, using the pretrained encoder-decoder language model BART \cite{lewis-etal-2020-bart} for conditional generation.
For each event type, we first initialize a type-specific prefix consisting of a sequence of tunable vectors as transformer history values \cite{li-liang-2021-prefix}.
The type-specific prefix offers tunable event type information for one single type.
Then we integrate context information with all type-specific prefixes to learn a context-specific prefix, dynamically combining all possible event type information.

We evaluate our model on two widely used event extraction benchmarks, ACE 2005 and ERE.
Experimental results show that our model achieves competitive results with the state-of-the-art classification-based model \textsc{OneIE} on ACE 2005 and achieves the best performances on ERE.
Additionally, according to the transfer learning results, our model also can be adapted to new types of events effectively.

\section{Related Work}

This paper is related to the following lines of work.

\subsection{Classification-based Event Extraction}

Event extraction is usually formulated as a sequence labeling classification problem \cite{nguyen-etal-2016-joint-event, wang-etal-2019-hmeae, yang-etal-2019-exploring, wadden-etal-2019-entity,liu-etal-2018-jointly}.
Some of them incorporate global features and apply joint inference \cite{lin-etal-2020-joint, li-etal-2013-joint, yang-mitchell-2016-joint} to collectively model event dependencies.
Additionally, recent work casts event extraction as a machine reading comprehension (MRC) problem \cite{liu-etal-2020-event, du-cardie-2020-event,li-etal-2020-event} by constructing questions to query event triggers and arguments.

Our work treats event extraction as a conditional generation task, which is more flexible and portable, which reduces the burden of annotation.

\subsection{Generation-based Event Extraction}

There is a rising line of work casting event extraction as a sequence generation problem, such as
transforming into translation tasks \cite{paolini2021structured}, generating with constrained decoding methods \cite{lu-etal-2021-text2event} and template-based conditional generation \cite{li-etal-2021-document,DBLP:journals/corr/abs-2108-12724}.

The two closest methods above \cite{li-etal-2021-document,DBLP:journals/corr/abs-2108-12724} both utilize manually designed discrete templates, which caused the sub-optimal problem.
Besides, the applied static type instruction does not consider the connections between events within the same context.
We replaced the static type instructions with the dynamic prefixes, which are continuous and tunable vectors during training, combining the manual event templates and alleviating the sub-optimal problem.

\begin{figure*}[!t]
  \centering
  \setlength{\belowcaptionskip}{-0.3cm}
  \includegraphics[width=118mm]{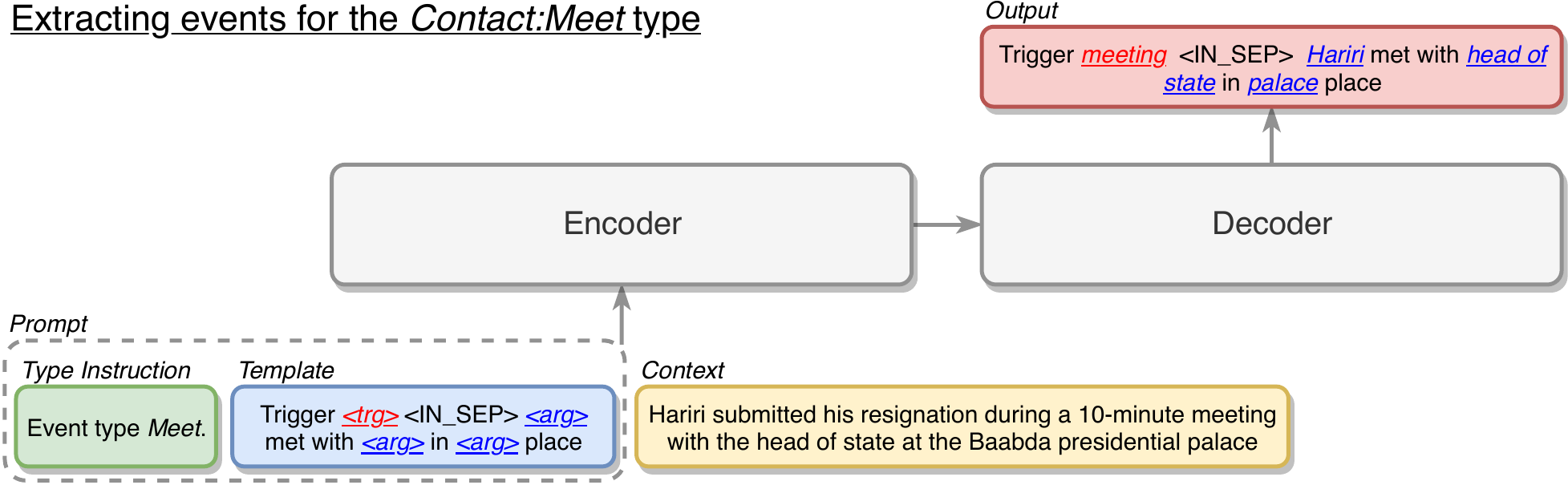}
  \caption{
    The framework of our base model \textsc{GTEE-Base}.
    We use different colors to differentiate different components as follows.
    `` \includegraphics[scale=0.2]{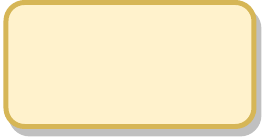} '' for the context,
    `` \includegraphics[scale=0.2]{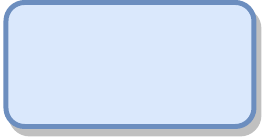} '' for the template,
    `` \includegraphics[scale=0.2]{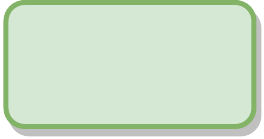} '' for the type instruction,
    `` \includegraphics[scale=0.2]{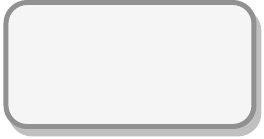} '' for the encoder-decoder language model,
    and `` \includegraphics[scale=0.2]{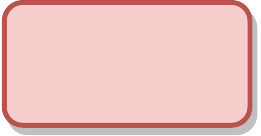} '' for the answered prompt as output.
  }
  \label{fig:base_model}
\end{figure*}

\subsection{Prompt Tuning}

There is a line of work using specific sentence templates with pre-trained models to solve natural language understanding tasks.
It natural to come up with prefix-style \cite{NEURIPS2020_1457c0d6} or cloze-style \cite{petroni-etal-2019-language} prompts based on human introspection, which are called ``descrete prompts''.
Existing works on discrete prompt tuning\cite{shin-etal-2020-autoprompt,gao-etal-2021-making, schick-etal-2020-automatically} depend on \textit{verbalizers} to map from class labels to answer tokens.
These methods are proven to be effective in the few-shot setting for text classification and conditional text generation tasks \cite{schick-schutze-2021-shot,schick-schutze-2021-exploiting,schick-schutze-2021-just}.
There are also methods that explore continuous prompts directly operating in the embedding space of the model, like tuning on vectors\cite{li-liang-2021-prefix,lester-etal-2021-power,DBLP:journals/corr/abs-2106-13884}, initializing with discrete prompts\cite{zhong-etal-2021-factual,qin-eisner-2021-learning,hambardzumyan-etal-2021-warp} and hybrid prompt tuning\cite{liu2021gpt,liu2021gptb,DBLP:journals/corr/abs-2105-11259}.

\section{Generative Template-based Method}

We revisit the task of event extraction as the process of conditional generation and present our base model (\textsc{GTEE-Base}) as illustrated in \cref{fig:base_model}.

\subsection{Problem Statement}

In the conditional generation task formulation for event extraction, the whole extraction process for a textual context is divided into several subtasks according to event types.
Specifically, given an event ontology $\mathcal{O}$ with an event type set $\mathcal{E}=\{e_i | i \in [1, |\mathcal{E}|]\}$, the input in each subtask $\mathcal{S}_{e_i, \mathcal{C}}$ for event type $e_i$ consists of a context $\mathcal{C}$ and a designed prompt $\mathcal{P}_{e_i}$.
And the output is the answered prompts $\mathcal{A}_{e_i}$, containing extracted event records.

We take one single conditional generation subtask $\mathcal{S}_{e_i, \mathcal{C}}$ for event type $e_i$ as example to explain the following content.

\subsection{Basic Architecture}

As shown in \cref{fig:base_model}, the conditional generation subtask is modeled by a pretrained encoder-decoder language model (LM), BART \cite{lewis-etal-2020-bart} and T5 \cite{DBLP:journals/jmlr/RaffelSRLNMZLL20}.
In the generation process, the encoder-decoder LM models the conditional probability of selecting a new token $y_i$ given the previous tokens $y_{<i}$ and the encoder input $\mathcal{X}$.
Therefore, the entire probability $p(\mathcal{Y} | \mathcal{X})$ of generating the output sequence $\mathcal{Y}$ given the input sequence $\mathcal{X}$ is calculated as
\begin{equation}
  \small
  \begin{split}
  p(\mathcal{Y} | \mathcal{X}) &= \prod_{i=1}^{|\mathcal{Y}|} p(y_i | y_{<i}, \mathcal{X}) \\
  \mathcal{X} &= [\mathcal{P}_{e_i}; \texttt{[SEP]}; \mathcal{C}] \\
  \mathcal{Y} &= \mathcal{A}_{e_i}
  \end{split}
  \label{eq:bart_cg}
\end{equation}
where $[\ ;\ ]$ denotes the sequence concatenation operation and \texttt{[SEP]}
\footnote{In this paper, we use \texttt{[*]} to represent the special tokens used in pretrained LM and \texttt{<*>} to indicate the user-defined special tokens.} is the corresponding separate marker in the applied encoder-decoder LM.

\subsection{Prompt Design}

Similar to the state-of-the-art end-to-end generative method \textsc{DEGREE-e2e} \cite{DBLP:journals/corr/abs-2108-12724} for event extraction, the prompt $\mathcal{P}_{e_i}$ for subtask $\mathcal{S}_{e_i, \mathcal{C}}$ in our base model \textsc{GTEE-Base} contains the type instruction $\mathcal{I}_{e_i}$ and the template $\mathcal{T}_{e_i}$.

\paragraph{Type Instruction.} A short natural language sequence $\mathcal{I}_{e_i}$ describing the event type $e_i$ in the subtask.
We use the pattern ``Event type is \texttt{[MASK]}.'' to construct type instructions for the event type set $\mathcal{E}$.
For example, the type instruction for event type \texttt{Meet} is ``Event type is Meet.''.

\paragraph{Template.} A type-specific pattern $\mathcal{T}_{e_i}$, which contains several placeholders, reflecting how the arguments participant in the event.
We use two types of placeholdes, \texttt{<trg>} and \texttt{<arg>}s, for representing trigger and arguments, respectively.
The template is consists of a trigger part and a argument part.
The two parts are concatenated by a new seperate marker \texttt{<IN\_SEP>}.
As illstrated in \cref{fig:base_model}, the trigger part is ``Trigger \texttt{<trg>}'', which is identical for all event types.
The argument part is specific to event type $e_i$.
Due to the manual efforts of designing and searching for an optimal template, we follow \citet{li-etal-2021-document} to reuse the pre-defined argument templates
\footnote{The argument template and all the used ontologies can be accessed at \url{https://github.com/raspberryice/gen-arg} except for ERE. Since the ERE event types are subsets of the RAMS AIDA ontology and the KAIROS ontology, following \citet{li-etal-2021-document}, we also reuse the argument templates from these ontologies.}
in the ontology $\mathcal{O}$.
The original pre-defined argument templates natively contain numeric labels for each \texttt{<arg>} placeholder (as \texttt{<arg1>}) and the slot mappings $\mathcal{M}$ to the corresponding argument roles.
We also follow \citet{li-etal-2021-document} to remove these numeric labels.

\paragraph{Ground Truth Construction.} For each event type $e_i$ in the context $\mathcal{C}$, we construct the ground truth sequence $\mathcal{G}_{e_i, \mathcal{C}}$ for conditional generation by filling the gold event records into the template $\mathcal{T}_{e_i}$.
If there is no event record of event type $e_i$, the generation ground truth will be ``Trigger \texttt{<trg>}''.
Otherwise, the event record is filled in the template $\mathcal{T}_{e_i}$ as the output in \cref{fig:base_model}.
If several arguments are categorized as the same role, these arguments are first sorted by spans and then concatenated by ``and''.
If there are multiple event records, they will be sorted by the spans of the triggers, and the filled sequences will be concatenated by a new separate marker \texttt{<OUT\_SEP>}.

\subsection{Training, Inference and Parsing}

\paragraph{Training.} The trainable parameters of our base model \textsc{GTEE-Base} are only the encoder-decoder LM.
And we use $\phi$ to denote all the trainable parameters.
Therefore, the training target is to minimize the negative loglikelihood of all subtasks $\mathcal{S}_{e_i, \mathcal{C}_j}$ in the training set $\mathcal{D}$, where $\mathcal{C}_j$ denotes the $j$-th context in $\mathcal{D}$.
\begin{equation}
  \small
  \begin{split}
  \mathcal{L}_{\phi}(\mathcal{D}) &= - \sum_{j=1}^{|\mathcal{D}|} \sum_{i=1}^{|\mathcal{E}|} \log p(\mathcal{G}_{e_i, \mathcal{C}_j} | \mathcal{X}_{e_i, \mathcal{C}_j}) \\
  \mathcal{X}_{e_i, \mathcal{C}_j} &= [\mathcal{P}_{e_i}; \texttt{[SEP]}; \mathcal{C}_j]
  \end{split}
  \label{eq:train_tgt}
\end{equation}

\paragraph{Inference.}
In the inference stage, our base model generates sequences by beam search $\texttt{BEAM}=6$.
The maximum sequence length is set according to dataset statistics, which is a bit larger than the length of the longest ground truth.

\paragraph{Parsing.}
Basically, we parse the event records by template matching and slot mapping according to the ontology $\mathcal{O}$.
Please note that not all the generated output sequences are valid.
For each generated sequence, we will first try to parse a trigger.
If failed, we will skip the sequence.
Then if we fail to match \texttt{<IN\_SEP>} or the argument part of the template $\mathcal{T}_{e_i}$, we will skip the argument parsing and only keep a trigger.

\begin{figure*}[!t]
  \centering
  \setlength{\belowcaptionskip}{-0.3cm}
  \includegraphics[width=\textwidth]{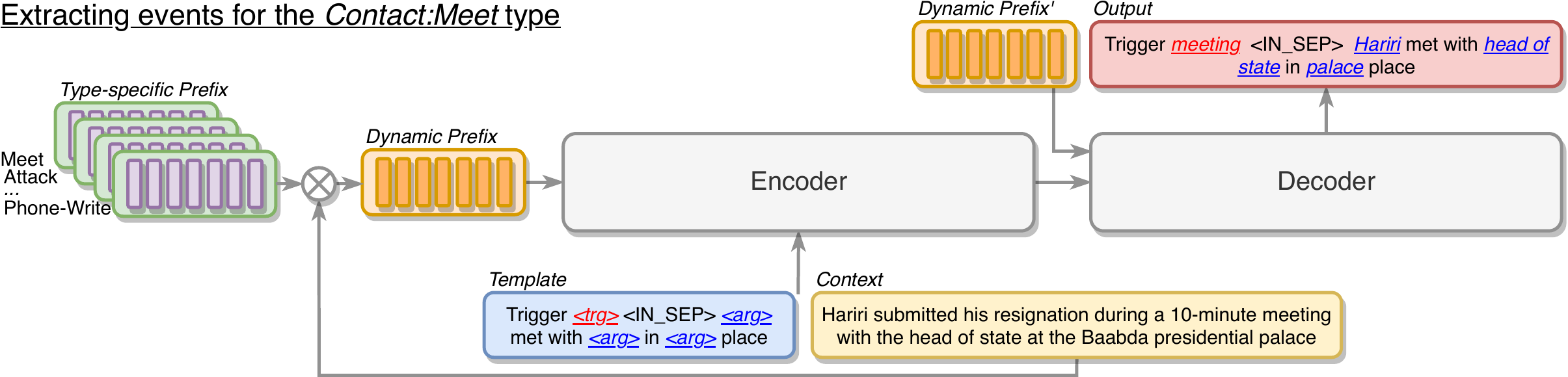}
  \caption{
    The framework of our dynamic prefix-tuning model \textsc{GTEE-DynPref}.
    We use different colors to differentiate different components as follows.
    `` \includegraphics[scale=0.2]{imgs/framework/context.crop.pdf} '' for the context,
    `` \includegraphics[scale=0.2]{imgs/framework/template.crop.pdf} '' for the template,
    `` \includegraphics[scale=0.2]{imgs/framework/type.crop.pdf} '' for the type-specific prefixes,
    `` \includegraphics[scale=0.2]{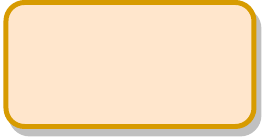} '' for the dynamic prefix,
    `` \includegraphics[scale=0.2]{imgs/framework/module.crop.pdf} '' for the encoder-decoder language model,
    and `` \includegraphics[scale=0.2]{imgs/framework/output.crop.pdf} '' for the answered prompt as output.
  }
  \label{fig:dp_model}
\end{figure*}

\subsection{Irrelevant Event Types}
\label{sec:negative_sample}

By investigating the parsed event records, we find that our model has the bias to generate event records even for irrelevant event types.
This will be fatal when the input context does not contain any event record, which will largely hurt the precision score and F1 score.
There are 80.28\% and 71.02\% sentences that do not contain any event records in ACE 2005 and ERE, respectively.

Therefore, we propose a simple yet effective solution to alleviate this problem by separately training an irrelevance classifier \textsc{IC}.
With context $\mathcal{C}$ as input, we finetune a \textsc{BERT} mdoel \cite{devlin-etal-2019-bert} by feeding the encoded \texttt{[CLS]} vector to a MLP as a binary classifier to see whether the context contains any event records or is entirely irrelevant for the ontology $\mathcal{O}$.
It is worth noticing that there may exist other ways to avoid the problem, as \citet{cui-etal-2021-template} formulate the NER task as a ranking task to avoid irrelevant entity types in a similar conditional generation task setting.

\section{Dynamic Prefix-Tuning}

We propose dynamic prefix-tuning with task-specific prefix and context-specific prefix to alleviate the two main challenges in generation-based event extraction.
The framework of our model with dynamic prefix tuning, \textsc{GTEE-DynPref}, is shown in \cref{fig:dp_model}.
We will introduce the dynamic prefix-tuning step by step.

\subsection{Type-Specific \textsc{Static Prefix}}

Inspired by \textsc{Prefix-Tuning} \cite{li-liang-2021-prefix}, we use event type-specific prefix \textsc{StaPref}, which is a pair of two transformer activation sequences $\{{sp}, {sp}^{\prime}\}$, each containing $L$ continuous $D$-dim vectors as the history values for the encoder and the decoder, respectively.
From the view of the encoder and decoder input, in the subtask $\mathcal{S}_{e_i, \mathcal{C}}$, the prefix is virtually prepended for the sequences $\mathcal{X}$ and $\mathcal{Y}$ in an encoder-decoder LM.
\begin{equation}
  \small
  \begin{split}
  \mathcal{X}^{\prime} &= [{sp}_{e_i}; \mathcal{X}] \\
  \mathcal{Y}^{\prime} &= [{sp}^{\prime}_{e_i}; \mathcal{Y}]
  \end{split}
\end{equation}
The main advantage of these transformer activation sequences is that they provide trainable context for both encoder and decoder, which is also computationally achievable.

We first initialize a pair of task-specific prefixes $\{{sp}_{e_i}, {sp}^{\prime}_{e_i}\}$ for each event type $e_i$ in the ontology $\mathcal{O}$.
In the conditional generation subtask $\mathcal{S}_{e_i, \mathcal{C}}$, we then prepend the corresponding pair of task-specific prefixes $\{{sp}_{e_i}, {sp}^{\prime}_{e_i}\}$ as transformer activations for the encoder and decoder.

Following \citet{li-liang-2021-prefix}, we use a trainable embedding tensor $P \in \mathbb{R}^{|\mathcal{E}| \times L \times D}$ to model the type-specific prefix $sp$.
For the event type $e_i$ in the ontology $\mathcal{O}$, the prefix vector $sp_{e_i}^{t}$ at index $t$ is
\begin{equation}
  \small
  sp_{e_i}^{t} = P[{e_i}, t, :]
\end{equation}

The reason we call the task-specific prefix \textit{static} is that for subtasks of the same event types, the output type instructions are the same.
In other words, such prefixes only preserve context concerning one single type of event, ignoring the association between different event types.

\begin{figure}[!t]
  \centering
  \setlength{\belowcaptionskip}{-0.3cm}
  \includegraphics[width=\columnwidth]{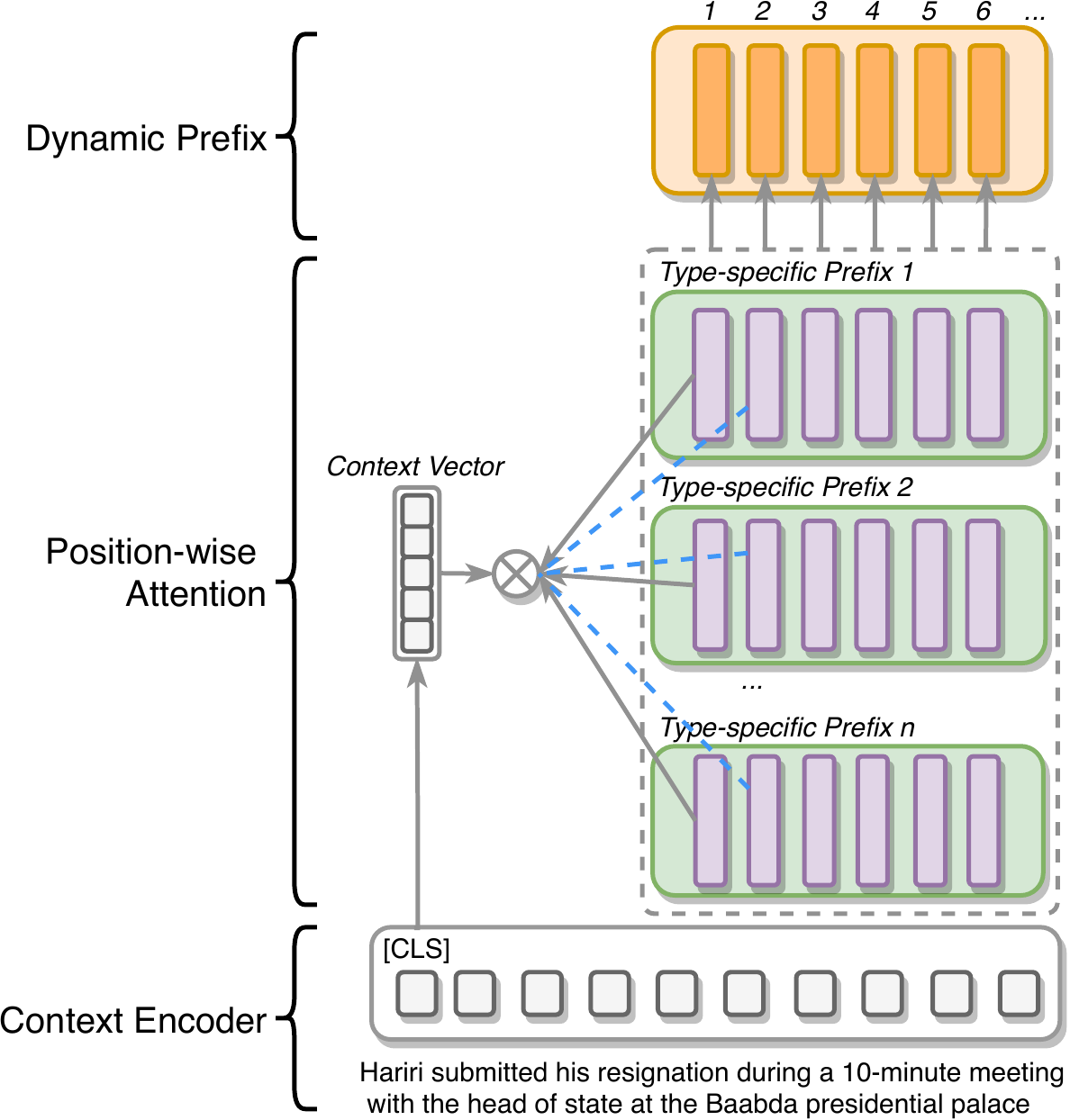}
  \caption{
    Context-specific \textsc{DynPref} construction using a context encoder.
  }
  \label{fig:dp_cal}
\end{figure}

\subsection{Context-Specific \textsc{Dynamic Prefix}}
\label{sec:dynpref}

Aiming to capture the associations between different event types when constructing trainable prefixes, we present \textsc{DynPref}, which constructs dynamic prefix with context-specific information when prompting pretrained language models.

As shown in \cref{fig:dp_cal}, ${dp}_{\mathcal{C}}$ has the same sequence length $L$ as ${sp}$.
For each position $t$, the prefix vector ${dp}_\mathcal{C}^{t}$ is computed by dynamically integrating all the prefix vector ${sp}_{e_i}^{t}$ of event type $e_i$ in the ontology $\mathcal{O}$ according to the context-specific information $c$ by multi-head attention \cite{DBLP:conf/nips/VaswaniSPUJGKP17}.
To calculate the context-specific information $c$, we apply a \textsc{BERT} mdoel \cite{devlin-etal-2019-bert} as the context encoder by feeding the context $\mathcal{C}$ as input and taking the \texttt{[CLS]} vector as $c$.
\begin{equation}
  \small
  \begin{split}
    {dp}_\mathcal{C}^{t} &= \mathop{\operatorname{MultiHeadAttn}}\limits_{i = 1}^{|\mathcal{E}|} (\{{sp}_{e_i}^{t},\ ...\}, c) \\
  c &= \operatorname{BERT}(\mathcal{C})_{\texttt{[CLS]}}
  \end{split}
\end{equation}

The context-specific prefix ${dp}_{\mathcal{C}}$ is \textit{dynamic} because it takes both the type-specific information in ontology $\mathcal{O}$ and the unique context information into account when steering LMs.

Following \citet{li-liang-2021-prefix}, we compute the decoder transformer activation vector $h_i$, which is a concatenation of all layers, at time step $i$ in encoder-decoder LM recurrently.
\begin{equation}
  \small
  h_{i} =
  \begin{cases}
  {dp}_{\mathcal{C}}^{i},  & \text{if } i < L \text{,} \\
  \operatorname{LM}(y_i, h_{<i} | \mathcal{X}),  & \text{otherwise.}
  \end{cases}
\end{equation}
The computation of the encoder transformer activation vector is similar.

\subsection{Training}

Except for the LM parameters $\phi$, the additional trainable parameters of \textsc{DynPref} include the embedding tensor $P$ and the \textsc{BERT} encoder modeling context information.

Specially, we follow the training suggestions \cite{li-liang-2021-prefix} and reparametrize the embedding tensor $P$ by modeling a \textsc{MLP} and another embedding tensor $P^{\prime} \in \mathbb{R}^{|\mathcal{E}| \times L \times D^{\prime}}$ with small dimension $D^{\prime} < D$.
In the end, $P$ is computed as
\begin{equation}
  \small
  P[{e_i}, t, :] = \operatorname{MLP}(P^{\prime}[{e_i}, t, :])
\end{equation}
Now we use $\theta$ to denote all the introduced parameters for \textsc{DynPref}.

The training objective is still to minimize the negative loglikelihood in \cref{eq:train_tgt} for $\phi$ and $\theta$.
However, in our preliminary experiments, we find that jointly learning the LM parameters $\phi$ and the \textsc{DynPref} parameters $\theta$ requires different scales of training hyperparameters, being difficult to learn the ability to extract event arguments.
Therefore, we train them separately in three steps:
(1) First, we train $\phi$ using \textsc{GTEE-Base} to learn the task information.
(2) Then we fix the LM parameters $\phi$ and mask all other event types except for $e_i$ in each subtask $\mathcal{S}_{e_i, \mathcal{C}}$, only optimizing $\theta$, to learn the type-specific information for each event type.
(3) Last, we remove the masking of event types, remaining the LM parameters fixed and only optimizing $\theta$ using \textsc{DynPref}, to capture the associations between related event types.

\section{Experiment Setup}

\subsection{Datasets}

We conducted experiments on two widely used event extraction benchmarks, ACE 2005 (LDC2006T06) and ERE (LDC2015E29, LDC2015E68, and LDC2015E78).
ACE 2005 dataset has 599 annotated English documents, 33 event types, and 22 argument roles.
ERE contains 458 English documents, 38 event types, and 21 argument roles.

We preprocess the datasets following previous work \cite{Zhang:2019:GAIL,wadden-etal-2019-entity,du-cardie-2020-event,lin-etal-2020-joint,lu-etal-2021-text2event,DBLP:journals/corr/abs-2108-12724}, and obtain three datasets, ACE05-E, ACE05-E$^+$ and ERE-EN.
Statistics of the datasets are shown in \cref{tab:dataset}.
Compared to ACE05-E, both ACE05-E$^+$ and ERE-EN contain pronoun roles and multi-token event triggers.

\begin{table}[t!]
\centering
\small
\setlength{\belowcaptionskip}{-0.3cm}
\resizebox{\columnwidth}{!}{
    \begin{tabular}{lcccc}
        \hline
        \textbf{Dataset}                   & \textbf{Split} & \textbf{\#Sents} & \textbf{\#Events} & \textbf{\#Roles} \\
        \hline
        \multirow{3}[2]{*}{ACE05-E}       & Train          & 17,172           & 4202             & 4859            \\
                                            & Dev            & 923              & 450               & 605              \\
                                            & Test           & 832              & 403               & 576              \\
        \multirow{3}[2]{*}{ACE05-E$^{+}$} & Train          & 19,216           & 4419             & 6607            \\
                                            & Dev            & 901              & 468               & 759              \\
                                            & Test           & 676              & 424               & 689              \\
        \multirow{3}[2]{*}{ERE-EN}         & Train          & 14,736           & 6208             & 8924            \\
                                            & Dev            & 1209            & 525               & 730              \\
                                            & Test           & 1163            & 551               & 822              \\
        \hline
    \end{tabular}
}
\caption{Dataset statistics.}
\label{tab:dataset}
\end{table}%

\subsection{Evaluation Metrics}

We use the same evaluation criteria in previous work \citep{Zhang:2019:GAIL,wadden-etal-2019-entity,lin-etal-2020-joint,lu-etal-2021-text2event,DBLP:journals/corr/abs-2108-12724} and report the Precision $P$, Recall $R$ and F1 score $F1$ of trigger classification (\textbf{Trg-C}) and argument classification (\textbf{Arg-C}).

\begin{itemize}
  \item \textbf{Trg-C}: a trigger is correctly classified if its offset and event type matches the ground truth.
  \item \textbf{Arg-C}: an argument is correctly classified if its offset, event type and role label all matches the ground truth.
\end{itemize}

Following \citet{lu-etal-2021-text2event}, we also obtain the offset of extracted triggers by string matching in the input context one by one.
For the predicted argument, we find the nearest matched string to the predicted trigger as the predicted offset.

\subsection{Baseline Methods}

We compare \textsc{GTEE-DynPref} with two groups of event extraction work.
The first group is about classification-based event extraction methods.
\begin{itemize}
  \item \textsc{DyGIE++} \cite{wadden-etal-2019-entity}: a BERT-based model which captures both within-sentence and cross-sentence context.
  \item \textsc{GAIL} \cite{Zhang:2019:GAIL}: an RL model jointly extracting entity and event.
  \item \textsc{OneIE} \cite{lin-etal-2020-joint}: an end-to-end IE system which employs global feature and beam search, which is the state-of-the-art.
  \item \textsc{BERT\_QA} \cite{du-cardie-2020-event}: a MRC-based model using multi-turns of separated QA pairs to extract triggers and arguments.
  \item \textsc{MQAEE} \cite{li-etal-2020-event}: a multi-turn question answering system.
\end{itemize}

The second group contains generation-based event extraction methods.
\begin{itemize}
  \item \textsc{TANL} \cite{paolini2021structured}: a method use translation tasks modeling event extraction in a trigger-argument pipeline.
  \item \textsc{BART-Gen} \cite{li-etal-2021-document}: a template-based conditional generation method.
  \item \textsc{Text2Event} \cite{lu-etal-2021-text2event}: a sequence-to-structure generation method.
  \item \textsc{DEGREE-e2e} \cite{DBLP:journals/corr/abs-2108-12724}: an end-to-end conditional genration method with discrete prompts.
\end{itemize}

\subsection{Implementation Details}

We use the huggingface implementation of \textsc{BART}-large as the encoder-decoder LM and \textsc{BERT}-large as the binary irrelevance classifier \textsc{IC} in \cref{sec:negative_sample} and the context encoder in \cref{sec:dynpref}.
We optimized our models by AdamW \citep{loshchilov2018decoupled}.
The hyperparameters we used are shown in \cref{tab:hyp}.
Each experiment is conducted on NVIDIA A100 Tensor Core GPU 40GB.
For simplicity, we randomly initialize\footnote{The random initialization is implemented in the \texttt{torch.nn.EmbeddingLayer} class in \texttt{PyTorch} v1.7.1.} the embedding tensor $P^{\prime}$.

As mentioned in \cref{sec:negative_sample}, there is an overwhelming amount of negative samples compared with positive samples.
Therefore, we sample only 4\% negative samples in the train and dev split for the three datasets, keeping all samples in test split.

\begin{table}[t!]
\centering
\small
\setlength{\belowcaptionskip}{-0.3cm}
\resizebox{\columnwidth}{!}{
    \begin{tabular}{lccc}
        \hline
        \textbf{Name} & \textsc{GTEE-Base} & \textsc{IC} & \textsc{GTEE-DynPref} \\
        \hline
        learning rate & 1e-5 & 2e-5 & 5e-5 \\
        train batch size & 32*8 & 16*8 & 32*8 \\
        epochs & 40 & 12 & 30 \\
        weight decay & 1e-5 & 1e-5 & 1e-5 \\
        gradient clip & 5.0 & 5.0 & 5.0 \\
        warm-up ratio & 10\% & 10\% & 10\% \\
        prefix length $L$ & - & - & 80 \\
        embedding dim $D^{\prime}$ & - & - & 512 \\
        \hline
    \end{tabular}
}
\caption{Hyperparameter setting for our models.}
\label{tab:hyp}
\end{table}%

\begin{table}[t!]
\small
\centering
\setlength{\belowcaptionskip}{-0.3cm}
\resizebox{\columnwidth}{!}{
    \begin{tabular}{lcccccc}
        \hline
        \multicolumn{1}{l}{\multirow{2}{*}{\textbf{Model}}} & \multicolumn{3}{c}{\textbf{Trg-C}} & \multicolumn{3}{c}{\textbf{Arg-C}} \\
        \multicolumn{1}{c}{} & P & R & F1 & P & R & F1 \\
        \hline
        \ \ \ \ \textit{classification-based} \\
        \textsc{DyGIE++} & - & - & 69.7 & - & - & 48.8 \\
        \textsc{GAIL} & 74.8 & 69.4 & 72.0 & 61.6 & 45.7 & 52.4 \\
        \textsc{OneIE} & - & - & \textbf{74.7} & - & - & \textbf{56.8} \\
        \textsc{BERT\_QA} & 71.1 & 73.7 & 72.3 & 56.8 & 50.2 & 53.3 \\
        \textsc{MQAEE} & - & - & 71.7 & - & - & 53.4 \\
        \hline
        \ \ \ \ \textit{generation-based} \\
        \textsc{TANL} & - & - & 68.5 & - & - & 48.5 \\
        \textsc{BART-Gen} & 69.5 & 72.8 & 71.1 & 56.0 & 51.6 & 53.7 \\
        \textsc{Text2Event} & 67.5 & 71.2 & 69.2 & 46.7 & 53.4 & 49.8 \\
        \textsc{DEGREE-e2e} & - & - & 70.9 & - & - & 54.4 \\
        \textbf{\textsc{GTEE-DynPref}} & 63.7 & 84.4 & \textbf{72.6} & 49.0 & 64.8 & \textbf{55.8} \\
        \hline
    \end{tabular}
    }
\caption{Results on ACE05-E for event extraction in the supervised learning setting. The first group of baselines is the classification-based methods and the second group is the generation-based methods.
Our proposed \textsc{GTEE-DynPref} is also the generation-based method.
For each group, we bold the highest F1 scores for Trg-C and Arg-C.}
\label{tab:supervised_learning1}
\end{table}%

\begin{table*}[t!]
\small
\centering
\setlength{\belowcaptionskip}{-0.3cm}
\resizebox{0.7\textwidth}{!}{
    \begin{tabular}{lcccccccccccc}
        \hline
        \multicolumn{1}{l}{\multirow{3}{*}{\textbf{Model}}} & \multicolumn{6}{c}{\textbf{ACE05-E$^{+}$}} & \multicolumn{6}{c}{\textbf{ERE-EN}} \\
        \multicolumn{1}{c}{} & \multicolumn{3}{c}{\textbf{Trg-C}} & \multicolumn{3}{c}{\textbf{Arg-C}} & \multicolumn{3}{c}{\textbf{Trg-C}} & \multicolumn{3}{c}{\textbf{Arg-C}} \\
        \multicolumn{1}{c}{} & P & R & F1 & P & R & F1 & P & R & F1 & P & R & F1 \\
        \hline
        \textsc{OneIE} & \textbf{72.1} & 73.6 & 72.8 & \textbf{55.4} & 54.3 & \textbf{54.8} & 58.4 & 59.9 & 59.1 & 51.8 & 49.2 & 50.5 \\
        \textsc{Text2Event} & 71.2 & 72.5 & 71.8 & 54.0 & 54.8 & 54.4 & 59.2 & 59.6 & 59.4 & 49.4 & 47.2 & 48.3 \\
        \textsc{DEGREE-e2e} & - & - & 72.7 & - & - & 55.0 & - & - & 57.1 & - & - & 49.6 \\
        \textbf{\textsc{GTEE-DynPref}} & 67.3 & \textbf{83.0} & \textbf{74.3} & 49.8 & \textbf{60.7} & 54.7 & \textbf{61.9} & \textbf{72.8} & \textbf{66.9} & \textbf{51.9} & \textbf{58.8} & \textbf{55.1} \\
        \hline
    \end{tabular}
}
\caption{Results on ACE05-E$^{+}$ and ERE-EN for event extraction in the supervised learning setting. For each column, we bold the highest score.}
\label{tab:supervised_learning2}
\end{table*}%

\section{Results}

\begin{table*}[t!]
\small
\centering
\setlength{\belowcaptionskip}{-0.3cm}
\resizebox{\textwidth}{!}{
    \begin{tabular}{clcccccccccccccccccc}
        \hline
        &\multicolumn{1}{l}{\multirow{3}{*}{\textbf{Model}}} & \multicolumn{6}{c}{\textbf{ACE05-E}} & \multicolumn{6}{c}{\textbf{ACE05-E$^{+}$}} & \multicolumn{6}{c}{\textbf{ERE-EN}} \\
        &\multicolumn{1}{c}{} & \multicolumn{3}{c}{\textbf{Trg-C}} & \multicolumn{3}{c}{\textbf{Arg-C}} & \multicolumn{3}{c}{\textbf{Trg-C}} & \multicolumn{3}{c}{\textbf{Arg-C}} & \multicolumn{3}{c}{\textbf{Trg-C}} & \multicolumn{3}{c}{\textbf{Arg-C}} \\
        &\multicolumn{1}{c}{} & P & R & F1 & P & R & F1 & P & R & F1 & P & R & F1 & P & R & F1 & P & R & F1 \\
        \hline
        \multicolumn{1}{l}{\multirow{3}{*}{$\bigg\Uparrow $}}&\textbf{\textsc{GTEE-DynPref}} & \textbf{63.7} & \textbf{84.4} & \textbf{72.6} & \textbf{49.0} & \textbf{64.8} & \textbf{55.8} & \textbf{67.3} & \textbf{83.0} & \textbf{74.3} & \textbf{49.8} & \textbf{60.7} & \textbf{54.7} & \textbf{61.9} & \textbf{72.8} & \textbf{66.9} & \textbf{51.9} & \textbf{58.8} & \textbf{55.1} \\
        \multicolumn{1}{c}{}&\textsc{GTEE-StaPref} & 62.8 & 83.9 & 71.8 & 47.0 & 64.2 & 54.3 & 66.5 & 82.8 & 73.7 & 49.1 & 60.4 & 54.2 & 61.4 & 72.2 & 66.4 & 50.7 & 58.5 & 54.3 \\
        \multicolumn{1}{c}{} &\textsc{GTEE-Base} & 61.9 & 83.4 & 71.0 & 46.4 & 63.7 & 53.7 & 65.7 & 82.1 & 73.0 & 48.1 & 59.7 & 53.2 & 60.6 & 71.3 & 65.5 & 49.8 & 57.8 & 53.5 \\
        \hline
    \end{tabular}
    }
\caption{Ablation study results on ACE05-E, ACE05-E$^{+}$ and ERE-EN.
From \textsc{GTEE-Base} to \textsc{GTEE-DynPref}, the model performances grows stronger.}
\label{tab:ablation}
\end{table*}%

\subsection{Supervised Learning Setting}

We evaluate the proposed model \textsc{GTEE-DynPref} under the supervised learning setting.
\cref{tab:supervised_learning1} shows the comparison results on ACE05-E against all baseline methods, and \cref{tab:supervised_learning2} illustrates the results compared with the state-of-the-art in each research line on ACE05-E$^{+}$ and ERE-EN.

\paragraph{New state-of-the-art.}
As we can see from \cref{tab:supervised_learning1}, \textsc{GTEE-DynPref} achieves the highest F1 scores for Trg-C and Arg-C on ACE05-E, compared with all the generation-based baselines.
Besides, \textsc{GTEE-DynPref} is competitive with the state-of-the-art classification-based method \textsc{OneIE}, outperforming the others.
In \cref{tab:supervised_learning2}, \textsc{GTEE-DynPref} achieves competitive Arg-C F1 score with \textsc{OneIE} on ACE05-E$^{+}$, while obtaining 7.5\% and 4.6\% gain of F1 scores for Trg-C and Arg-C, respectively, achieving new state-of-the-art on ERE-EN.

\paragraph{Trainable prompts boost the performances.}
Compared with \textsc{DEGREE}, the event extraction method using fixed templates, and \textsc{Text2Event}, the generative event extraction method without prompts, \textsc{GTEE-DynPref} outperforms them in all the datasets, showing the effectiveness of the trainable dynamic prefix with prompts.

\begin{table}[t!]
\small
\centering
\setlength{\belowcaptionskip}{-0.3cm}
\resizebox{\columnwidth}{!}{
    \begin{tabular}{lcccccc}
        \hline
        \multicolumn{1}{l}{\multirow{2}{*}{\textbf{Model}}} & \multicolumn{3}{c}{\textbf{Trg-C}} & \multicolumn{3}{c}{\textbf{Arg-C}} \\
        \multicolumn{1}{c}{} & P & R & F1 & P & R & F1 \\
        \hline
        \textsc{OneIE} w/o TL & 70.8 & 64.8 & 67.7 & 53.2 & 37.5 & 44.0 \\
        \textsc{OneIE} w/ TL & 71.0 & 64.4 & 67.6 & 54.7 & 38.1 & 45.0 \\
        \ \ \ \ $\Delta$\textit{performance} & +0.2 & -0.4 & -0.1 & +1.5 & +0.6 & +1.0 \\
        \hline
        \textsc{Text2Event} w/o TL & 72.9 & 62.7 & 67.4 & 54.0 & 38.1 & 44.7 \\
        \textsc{Text2Event} w/ TL & 75.1 & 64.0 & 69.1 & 56.0 & 40.5 & 47.0 \\
        \ \ \ \ $\Delta$\textit{performance} & +2.2 & +1.3 & +1.7 & +2.0 & +2.4 & +2.3 \\
        \hline
        \textsc{GTEE-DynPref} w/o TL & 62.0 & 75.4 & 68.1 & 39.6 & 53.5 & 45.5 \\
        \textsc{GTEE-DynPref} w/ TL & 64.6 & 76.7 & 70.2 & 43.7 & 54.1 & 48.3 \\
        \ \ \ \ $\Delta$\textit{performance} & +2.6 & +1.3 & +2.1 & +4.1 & +0.6 & +2.8 \\
        \hline
    \end{tabular}
    }
\caption{Transfer learning results on ACE05-E$^{+}$.}
\label{tab:transfer_learning}
\end{table}%

\subsection{Transfer Learning Setting}

\textsc{GTEE-DynPref} utilizes the event type templates and optimize them with context-specific information in the dynamic prefix, which is easy to extend to a new type of event.
Therefore, aiming to verify the ability of \textsc{GTEE-DynPref} to learn from new event types, we conduct experiments under the transfer learning setting following \citet{lu-etal-2021-text2event}.
Specifically, we divide the event mentions whose context contains no less than eight tokens in ACE05-E$^{+}$ into two subsets, denoted by \texttt{src} and \texttt{tgt}.
\texttt{src} contains top-10 frequent types of events and \texttt{tgt} contains the rest 23 types of events.
We then randomly split each subset into a \texttt{train} set and a \texttt{test} set with the ratio $4:1$.
Specifically, for transfer learning, we will first pre-train on \texttt{src-train} to learn the task information and then fine-tune on \texttt{tgt-train} for extracting the new types of events.
\cref{tab:transfer_learning} shows the evaluation results on \texttt{tgt-test} under the transfering learning setting and when solely training on \texttt{tgt-train} without transfering knowledge.
We choose the state-of-the-art classification-based model \textsc{OneIE} and generation-based method \textsc{Text2Event} as the baselines.

We can see that \textsc{GTEE-DynPref} achieves the highest Trg-C F1 and Arg-C F1 scores, which indicates that with the help of dynamic prefix, \textsc{GTEE-DynPref} can be adopted to new types of events more effectively.
Additionally, comparing with solely training on \texttt{tgt}, transfering the knowledge from \texttt{src} allows \textsc{GTEE-DynPref} to achieve higher F1 scores than \textsc{OneIE} and \textsc{Text2Event}.
The reason may be that \textsc{OneIE} relies on multi-task annotated information, and \textsc{Text2Event} requires learning the structural information of new types of events.
In contrast, \textsc{GTEE-DynPref} only requires an easy-to-acquire template, which can be further optimized during training.

\begin{table}[t!]
\small
\centering
\setlength{\belowcaptionskip}{-0.3cm}
\resizebox{\columnwidth}{!}{
    \begin{tabular}{lcccccc}
        \hline
        \multicolumn{1}{l}{\multirow{2}{*}{\textbf{Model}}} & \multicolumn{2}{c}{\textbf{ACE05-E}} & \multicolumn{2}{c}{\textbf{ACE05-E$^{+}$}} & \multicolumn{2}{c}{\textbf{ERE-EN}} \\
        \multicolumn{1}{c}{} & \textbf{Trg-C} & \textbf{Arg-C} & \textbf{Trg-C} & \textbf{Arg-C} & \textbf{Trg-C} & \textbf{Arg-C} \\
        \hline
        \textsc{GTEE-DynPref} \\
        \ \ \ \ w/o IC & 57.2 & 43.8 & 61.7 & 46.4 & 52.1 & 44.7 \\
        \ \ \ \ w/ IC (trained) & 72.6 & 55.8 & 74.3 & 54.7 & 66.9 & 55.1 \\
        \ \ \ \ w/ IC (gold) & 76.3 & 58.4 & 77.2 & 56.9 & 72.3 & 57.4 \\
        \hline
        \textsc{GTEE-StaPref} \\
        \ \ \ \ w/o IC & 56.9 & 43.4 & 61.3 & 45.9 & 51.4 & 44.0 \\
        \ \ \ \ w/ IC (trained) & 71.8 & 54.3 & 73.7 & 54.2 & 66.4 & 54.3 \\
        \ \ \ \ w/ IC (gold) & 75.2 & 57.5 & 76.6 & 55.8 & 71.6 & 56.9 \\
        \hline
        \textsc{GTEE-Base} \\
        \ \ \ \ w/o IC & 56.4 & 42.8 & 60.8 & 45.1 & 50.7 & 43.1 \\
        \ \ \ \ w/ IC (trained) & 71.0 & 53.7 & 73.0 & 53.2 & 65.5 & 53.5 \\
        \ \ \ \ w/ IC (gold) & 74.6 & 55.9 & 75.1 & 54.8 & 70.7 & 56.5 \\
        \hline
    \end{tabular}
    }
\caption{
The F1 scores under different irrelevance classifier settings on ACE05-E, ACE05-E$^{+}$ and ERE-EN.}
\label{tab:classifier}
\end{table}%

\subsection{Ablation Study}

In this section, we study the effectiveness of each proposed module by adding them into our base model \textsc{GTEE-Base} and finally get our final model \textsc{GTEE-DynPref}.
The results on ACE05-E, ACE05-E$^{+}$ and ERE-EN are presented in \cref{tab:ablation}.

\paragraph{Continuous Prompt vs Discrete Prompt.}
We first compare \textsc{GTEE-StaPref} with \textsc{GTEE-Base}.
Based on \textsc{GTEE-Base} with discrete prompts, \textsc{GTEE-StaPref} further combines type-specific prefixes as to form continuous prompts.
It can be observed that there is a 0.8\%, 0.7\% and 0.9\% gain for the Trg-C F1 score on ACE05-E, ACE05-E$^{+}$ and ERE-EN, respectively.
Additionally, there is a 0.6\%, 1.0\% and 0.8\% improvement for the Arg-C F1 score, demonstrating the effectiveness and flexibility of \textsc{StaPref} to model the type-specific information.

\paragraph{Dynamic Prefix vs Static Prefix.}
Next we compare \textsc{GTEE-DynPref} with \textsc{GTEE-StaPref} to study the advantages of constructing dynamic prefix.
On the basis of \textsc{GTEE-StaPref}, integrating context-specific information leads to a constent gain for Trg-C F1 score on all the datasets as 0.8\%, 0.6\% and 0.5\%, respectively.
There can also be observed a 1.5\%, 0.5\% and 0.8\% increase for the Arg-C F1 scores, respectively.
It indicates that integrating context-specific information into type-specific information and transforming static prefix to dynamic is beneficial for generative template-based event extraction.

\subsection{Irrelevance Classifier}

Our goal of the irrelevance classifier \textsc{IC} is to recognize the context that does not contain any event records in a given ontology $\mathcal{O}$.
According to \cref{sec:negative_sample}, we train an IC and use it for each dataset separately.
Please note that on one specific dataset, we will use the same IC for all the experiments corresponding to that dataset.
The accuracy of \textsc{IC} is 95.4\%, 93.5\% and 94.2\% for ACE05E, ACE05E$^{+}$ and ERE-EN, respectively.
To further study the influence of IC, we compare the performances of using no IC, trained IC, and gold IC.
The compared F1 scores are listed in \cref{tab:classifier}.

First, we find that with the help of our trained ICs on each dataset, the Trg-C and Arg-C F1 scores have been improved a lot by more than ten percentage points, indicating the necessity of IC.
Second, by replacing the trained IC with the oracle gold IC results, we can still observe possible increasements for F1 scores, suggesting the existence of likely chances for further optimizing IC performances.
We leave the optimization for IC as future work.

\begin{figure}[!tpb]
\centering
\setlength{\belowcaptionskip}{-0.3cm}
\subfigure[F1 scores of \textsc{GTEE-DynPref} with different prefix length $L$.]{
    \resizebox{0.8\columnwidth}{!}{
    \includegraphics[width=0.8\columnwidth]{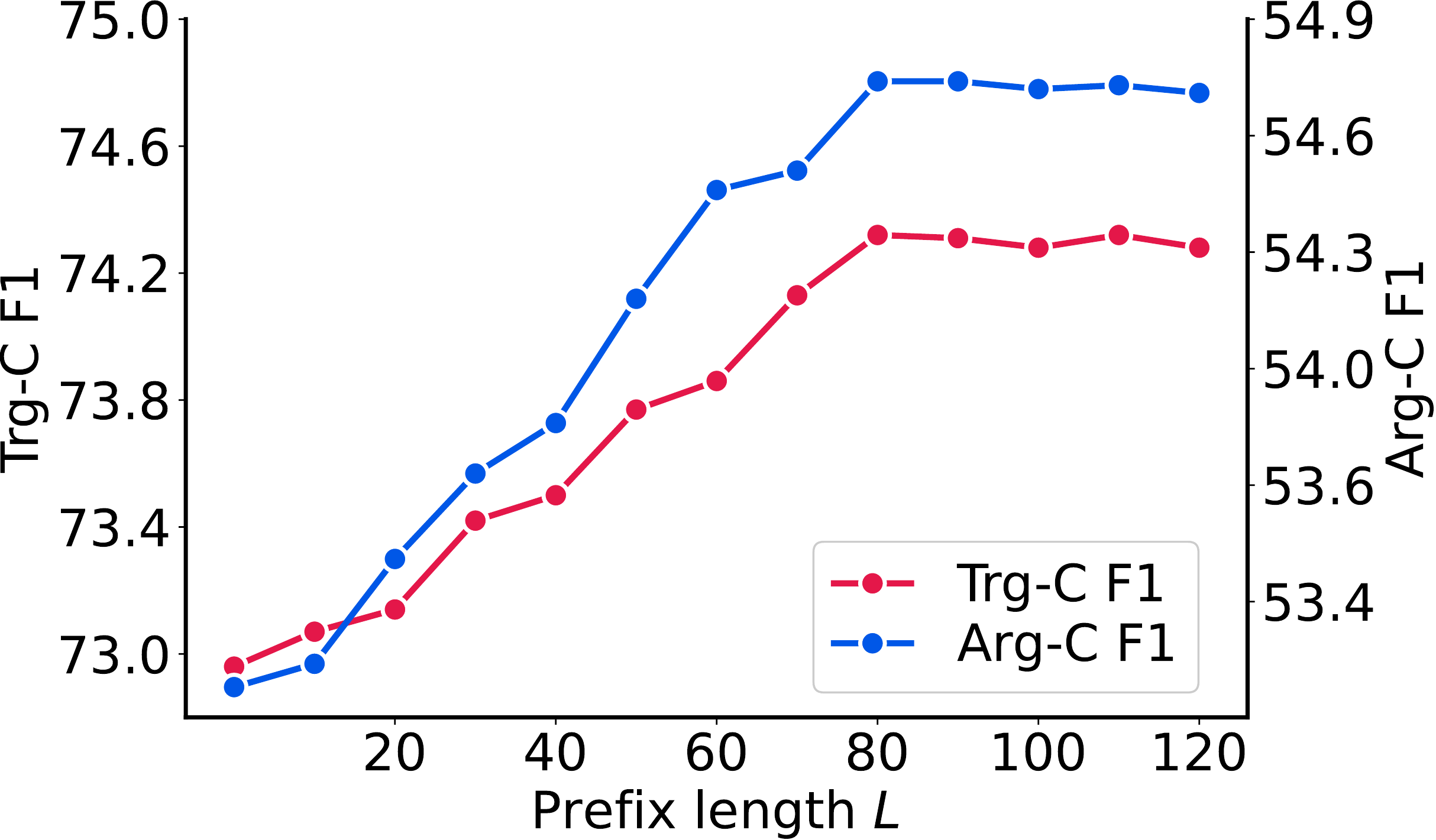}
    \label{fig:length}
    }
}
\subfigure[Performances of \textsc{GTEE-DynPref} with different dimension $D^{\prime}$ of the embedding tensor $P^{\prime}$.]{
    \resizebox{0.8\columnwidth}{!}{
    \includegraphics[width=0.8\columnwidth]{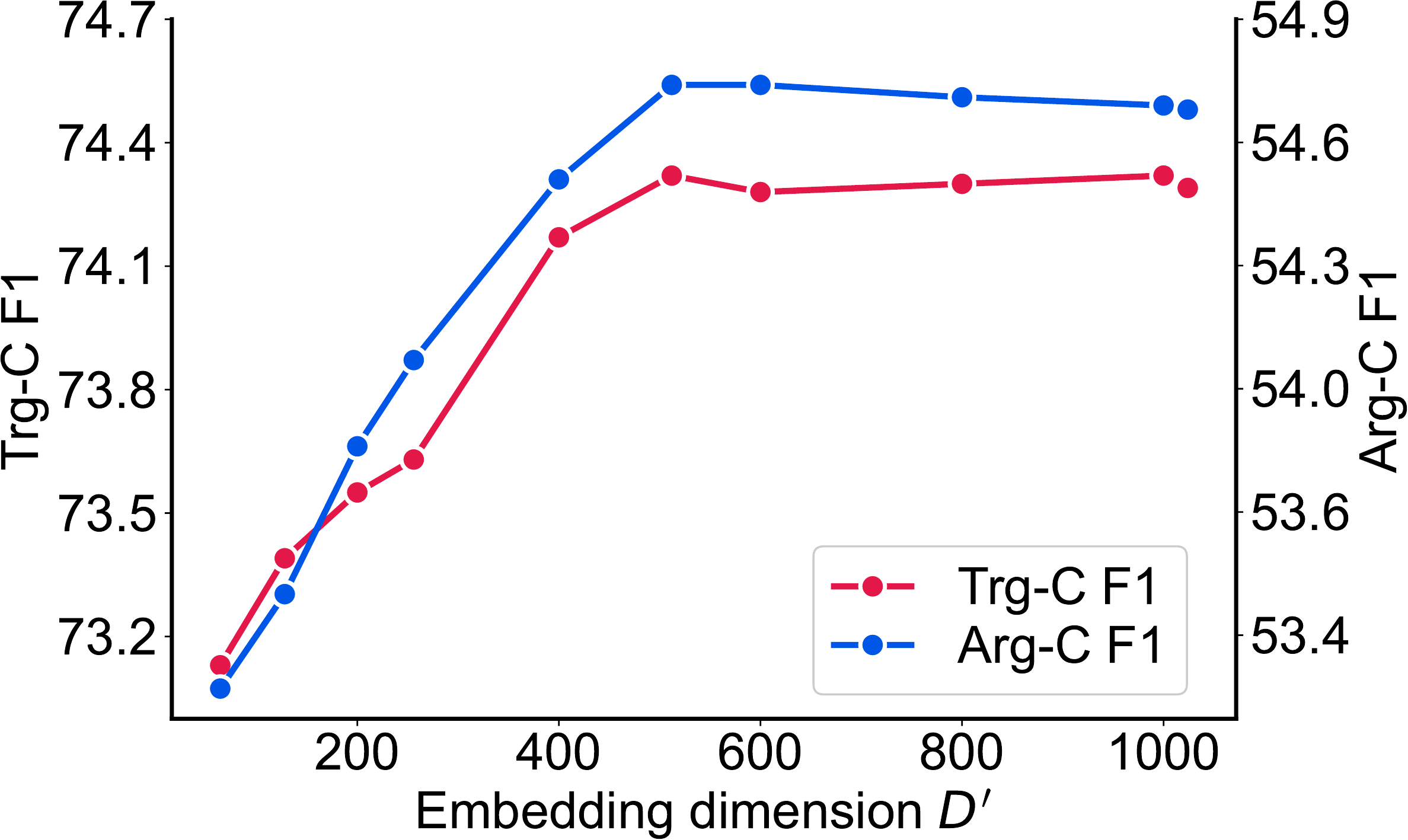}
    \label{fig:dim}
    }
}
\caption{Intrinsic evaluation results on ACE05-E$^{+}$.}
\label{fig:prefix}
\end{figure}%

\subsection{Intrinsic Evaluation}

We study the intrinsic characteristics of \textsc{GTEE-DynPref} by showing the influences of model hyperparameters on ACE05-E$^{+}$.

\paragraph{Prefix length $L$.}
We first study the impact of prefix length $L$ by grid search in $\{L | L=10*k, k \in \mathbb{N} \land k \le 12\}$.
\cref{fig:length} shows the Trg-C and Arg-C F1 scores.
We can observe that both Trg-C and Arg-C F1 scores increase as the prefix length $L$ increases to 80, afterward, a slight fluctuation.
We think the longer $L$ introduces more trainable parameters and a more vital ability to model the context-specific type information.
Therefore, we choose 80 as the prefix length in \textsc{GTEE-DynPref}.

\paragraph{Embedding dimension $D^{\prime}$.}
Similarly, we study the impact of the dimension $D^{\prime}$ of the embedding tensor $P^{\prime}$ by increasing from 64 to 1024.
The results of Trg-C and Arg-C F1 scores are illustrated in \cref{fig:dim}.
We find that although the bigger embedding dimension $D^{\prime}$ theoretically provides expressive type-specific information and improves the F1 scores when $D^{\prime} <= 512$, the continual improvement is interrupted when the embedding dimension is around 512.
Thus we set the embedding dimension $D^{\prime}=512$ in \textsc{GTEE-DynPref}.

\section{Conclusion}

In this paper, we studied event extraction in the template-based conditional generation manner.
We proposed the dynamic prefix tuning model \textsc{GTEE-DynPref} for event extraction.
On the one hand the method constructs tunable prefixes to model type-specific information and on the other hand \textsc{GTEE-DynPref} captures the associations between event types and calculates a context-specific prefix when steering pretrained language models.
Experimental results show that our model achieves competitive results with the state-of-the-art on ACE 2005, which is also proven to be portable to new types of events effectively.

\section{Ethical Consideration}

Event extraction is a standard task in NLP.
We do not see any significant ethical concerns.
Our work is easy to adapt to new event types by offering some examples and pre-defined templates.
Therefore, the expected usages of our work is to identify interesting event records from user textual input such as a piece of sentence or document.

\section*{Acknowledgments}
We thank the anonymous reviewers for their valuable comments and suggestions.
This work is supported by National Natural Science Foundation of China (No. U19B2020 and No. 62106010).

\bibliography{custom}
\bibliographystyle{acl_natbib}

\appendix

\section{Argument Template}

We use templates for ACE and ERE.
\cref{tab:ace_template} and \cref{tab:ere_template} show the argument templates for ACE and ERE, respectively, which is from the RAMS AIDA ontology and the KAIROS ontology.

\begin{table*}[t!]
\small
\centering
\setlength{\belowcaptionskip}{-0.3cm}
\resizebox{\textwidth}{!}{
    \begin{tabular}{llccccc}
    \hline
    \textbf{Event Type} & \textbf{Template} & \textbf{arg1} & \textbf{arg2} & \textbf{arg3} & \textbf{arg4} & \textbf{arg5} \\
        \hline
Movement:Transport      & \makecell[l]{<arg1> transported <arg2> in <arg3> vehicle from\\ <arg4> place to <arg5> place }     &Agent      &Artifact       &Vehicle        &Origin     &Destination\\

Personnel:Elect     &<arg1> elected <arg2> in <arg3> place      &Entity     &Person     &Place      &-      &-\\

Personnel:Start-Position        &<arg1> started working at <arg2> organization in <arg3> place      &Person     &Entity     &Place      &-      &-\\

Personnel:Nominate      &<arg1> nominated <arg2>        &Agent      &Person     &-      &-      &-\\

Personnel:End-Position      &<arg1> stopped working at <arg2> organization in <arg3> place      &Person     &Entity     &Place      &-      &-\\

Conflict:Attack     &\makecell[l]{<arg1> attacked <arg2> hurting <arg5> victims \\ using <arg3> instrument at <arg4> place  }    &Attacker       &Target     &Instrument     &Place      &Victim\\

Contact:Meet        &<arg1> met with <arg2> in <arg3> place     &Entity     &Entity     &Place      &-      &-\\

Life:Marry      &<arg1> married <arg2> in <arg3> place      &Person     &Person     &Place      &-      &-\\

Transaction:Transfer-Money      &\makecell[l]{<arg1> gave money to <arg2> for \\ the benefit of <arg3> in <arg4> place }     &Giver      &Recipient      &Beneficiary        &Place      &-\\

Conflict:Demonstrate        &<arg1> demonstrated at <arg2> place        &Entity     &Place      &-      &-      &-\\

Business:End-Org        &<arg1> organization shut down at <arg2> place      &Org        &Place      &-      &-      &-\\

Justice:Sue     &<arg1> sued <arg2> before <arg3> court or judge in <arg4> place        &Plaintiff      &Defendant      &Adjudicator        &Place      &-\\

Life:Injure     &<arg1> injured <arg2> with <arg3> instrument in <arg4> place       &Agent      &Victim     &Instrument     &Place      &-\\

Life:Die        &<arg1> killed <arg2> with <arg3> instrument in <arg4> place        &Agent      &Victim     &Instrument     &Place      &-\\

Justice:Arrest-Jail     &<arg1> arrested <arg2> in <arg3> place     &Agent      &Person     &Place      &-      &-\\

Contact:Phone-Write     &<arg1> communicated remotely with <arg2> at <arg3> place       &Entity     &Entity     &Place      &-      &-\\

Transaction:Transfer-Ownership      &\makecell[l]{ <arg1> gave <arg4> to <arg2> for \\ the benefit of <arg3> at <arg5> place }    &Seller     &Buyer      &Beneficiary        &Artifact       &Place\\

Business:Start-Org      &<arg1> started <arg2> organization at <arg3> place     &Agent      &Org        &Place      &-      &-\\

Justice:Execute     &<arg1> executed <arg2> at <arg3> place     &Agent      &Person     &Place      &-      &-\\

Justice:Trial-Hearing       &<arg1> tried <arg2> before <arg3> court or judge in <arg4> place       &Prosecutor     &Defendant      &Adjudicator        &Place      &-\\

Life:Be-Born        &<arg1> was born in <arg2> place        &Person     &Place      &-      &-      &-\\

Justice:Charge-Indict       &\makecell[l]{ <arg1> charged or indicted <arg2> before \\<arg3> court or judge in <arg4> place }    &Prosecutor     &Defendant      &Adjudicator        &Place      &-\\

Justice:Convict     &<arg1> court or judge convicted <arg2> in <arg3> place     &Adjudicator        &Defendant      &Place      &-      &-\\

Justice:Sentence        &<arg1> court or judge sentenced <arg2> in <arg3> place     &Adjudicator        &Defendant      &Place      &-      &-\\

Business:Declare-Bankruptcy     &<arg1> declared bankruptcy at <arg2> place     &Org        &Place      &-      &-      &-\\

Justice:Release-Parole      &<arg1> released or paroled <arg2> in <arg3> place      &Entity     &Person     &Place      &-      &-\\

Justice:Fine        &<arg1> court or judge fined <arg2> at <arg3> place     &Adjudicator        &Entity     &Place      &-      &-\\

Justice:Pardon      &<arg1> court or judge pardoned <arg2> at <arg3> place      &Adjudicator        &Defendant      &Place      &-      &-\\

Justice:Appeal      &<arg1> appealed to <arg2> court or judge at <arg3> place       &Plaintiff      &Adjudicator        &Place      &-      &-\\

Justice:Extradite       &<arg1> extradited <arg2> from <arg3> place to <arg4> place     &Agent      &Person     &Origin     &Destination        &-\\

Life:Divorce        &<arg1> divorced <arg2> in <arg3> place     &Person     &Person     &Place      &-      &-\\

Business:Merge-Org      &<arg1> organization merged with <arg2> organization        &Org        &Org        &-      &-      &-\\

Justice:Acquit      &<arg1> court or judge acquitted <arg2>     &Adjudicator        &Defendant      &-      &-      &-\\
        \hline
    \end{tabular}
}
\caption{All argument templates for ACE05-E and ACE05-E$^{+}$.}
\label{tab:ace_template}
\end{table*}%

\begin{table*}[t!]
\small
\centering
\setlength{\belowcaptionskip}{-0.3cm}
\resizebox{\textwidth}{!}{
    \begin{tabular}{llccccc}
    \hline
    \textbf{Event Type} & \textbf{Template} & \textbf{arg1} & \textbf{arg2} & \textbf{arg3} & \textbf{arg4} & \textbf{arg5} \\
        \hline
Conflict:Attack		&<arg1> attacked <arg2> using <arg3> instrument at <arg4> place		&Attacker		&Target		&Instrument		&Place		&-\\

Justice:Acquit		&<arg1> court or judge acquitted <arg2> at <arg3> place		&Adjudicator		&Defendant		&Place		&-		&-\\

Personnel:Elect		&<arg1> elected <arg2> in <arg3> place		&Agent		&Person		&Place		&-		&-\\

Justice:Release-Parole		&<arg1> released or paroled <arg2> in <arg3> place		&Agent		&Person		&Place		&-		&-\\

Personnel:Nominate		&<arg1> nominated <arg2> at <arg3> place		&Agent		&Person		&Place		&-		&-\\

Justice:Appeal		&<arg1> appealed to <arg2> court or judge sentenced <arg3>		&Prosecutor		&Adjudicator		&Defendant		&-		&-\\

Transaction:Transfer-Ownership		&\makecell[l]{<arg1> gave <arg4> to <arg2> for \\ the benefit of <arg3> at <arg5> place}		&Giver		&Recipient		&Beneficiary		&Thing		&Place\\

Business:Declare-Bankruptcy		&<arg1> declared bankruptcy		&Org		&-		&-		&-		&-\\

Contact:Meet		&<arg1> met face-to-face with <arg2> in <arg3> place		&Entity		&Entity		&Place		&-		&-\\

Life:Marry		&<arg1> married <arg2> in <arg3> place		&Person		&Person		&Place		&-		&-\\

Life:Divorce		&<arg1> divorced <arg2> in <arg3> place		&Person		&Person		&Place		&-		&-\\

Business:Merge-Org		&<arg1> organization merged with <arg2> organization		&Org		&Org		&-		&-		&-\\

Contact:Correspondence		&<arg1> communicated remotely with <arg2> at <arg3> place		&Entity		&Entity		&Place		&-		&-\\

Contact:Contact		&<arg1> communicated with <arg2> at <arg3> place		&Entity		&Entity		&Place		&-		&-\\

Manufacture:Artifact		&<arg1> manufactured or created or produced <arg2> at <arg3> place		&Agent		&Artifact		&Place		&-		&-\\

Movement:Transport-Person		&\makecell[l]{ <arg1> transported <arg2> in <arg3> instrument \\ from <arg4> place to <arg5> place}		&Agent		&Person		&Instrument		&Origin		&Destination\\

Movement:Transport-Artifact		&<arg1> transported <arg2> from <arg3> place to <arg4> place		&Agent		&Artifact		&Origin		&Destination		&-\\

Contact:Broadcast		&\makecell[l]{ <arg1> communicated to <arg2> at <arg3> place\\ (one-way communication) }		&Entity		&Audience		&Place		&-		&-\\

Transaction:Transaction		&\makecell[l]{<arg1> gave something to <arg2> for \\ the benefit of <arg3> at <arg4> place	}	&Giver		&Recipient		&Beneficiary		&Place		&-\\

Personnel:Start-Position		&<arg1> started working at <arg2> organization in <arg3> place		&Person		&Entity		&Place		&-		&-\\

Justice:Pardon		&<arg1> court or judge pardoned <arg2> at <arg3> place		&Adjudicator		&Defendant		&Place		&-		&-\\

Justice:Fine		&<arg1> court or judge fined <arg2> at <arg3> place		&Adjudicator		&Entity		&Place		&-		&-\\

Justice:Trial-Hearing		&\makecell[l]{<arg1> tried <arg2> before <arg3> \\ court or judge in <arg4> place }		&Prosecutor		&Defendant		&Adjudicator		&Place		&-\\

Business:End-Org		&<arg1> organization shut down at <arg2> place		&Org		&Place		&-		&-		&-\\

Justice:Sue		&\makecell[l]{ <arg1> sued <arg2> before <arg3> court or judge  \\ in <arg4> place }		&Plaintiff		&Defendant		&Adjudicator		&Place		&-\\

Life:Injure		&<arg1> injured <arg2> with <arg3> instrument in <arg4> place		&Agent		&Victim		&Instrument		&Place		&-\\

Justice:Arrest-Jail		&<arg1> arrested <arg2> in <arg3> place		&Agent		&Person		&Place		&-		&-\\

Justice:Execute		&<arg1> executed <arg2> at <arg3> place		&Agent		&Person		&Place		&-		&-\\

Conflict:Demonstrate		&<arg1> demonstrated at <arg2> place		&Entity		&Place		&-		&-		&-\\

Justice:Sentence		&<arg1> court or judge sentenced <arg2> in <arg3> place		&Adjudicator		&Defendant		&Place		&-		&-\\

Life:Die		&<arg1> killed <arg2> with <arg3> instrument in <arg4> place		&Agent		&Victim		&Instrument		&Place		&-\\

Business:Start-Org		&<arg1> started <arg2> organization at <arg3> place		&Agent		&Org		&Place		&-		&-\\

Personnel:End-Position		&<arg1> stopped working at <arg2> organization in <arg3> place		&Person		&Entity		&Place		&-		&-\\

Justice:Extradite		&<arg1> extradited <arg2> from <arg3> place to <arg4> place		&Agent		&Person		&Origin		&Destination		&-\\

Justice:Charge-Indict		&\makecell[l]{ <arg1> charged or indicted <arg2> before <arg3>  \\ court or judge in <arg4> place }		&Prosecutor		&Defendant		&Adjudicator		&Place		&-\\

Transaction:Transfer-Money		&\makecell[l]{ <arg1> gave money to <arg2> for  \\ the benefit of <arg3> in <arg4> place}		&Giver		&Recipient		&Beneficiary		&Place		&-\\

Justice:Convict		&<arg1> court or judge convicted <arg2> in <arg3> place		&Adjudicator		&Defendant		&Place		&-		&-\\

Life:Be-Born		&<arg1> was born in <arg2> place		&Person		&Place		&-		&-		&-\\
        \hline
    \end{tabular}
}
\caption{All argument templates for ERE-EN.}
\label{tab:ere_template}
\end{table*}%

\section{Transfer Learning Details}

The top-10 frequent types of events in the \texttt{src} split of ACE05-E$^{+}$ are listed as follows:
\begin{itemize}
  \item Transaction:Transfer-Ownership
  \item Contact:Phone-Write
  \item Personnel:Elect
  \item Personnel:End-Position
  \item Movement:Transport
  \item Life:Injure
  \item Conflict:Attack
  \item Transaction:Transfer-Money
  \item Contact:Meet
  \item Life:Die
\end{itemize}

\end{document}